\documentclass[twoside,11pt]{article}

\usepackage{jmlr2e}

\usepackage{amsmath}
\usepackage{cases}

\usepackage{algorithm}
\usepackage{algpseudocode}

\usepackage{graphicx}
\graphicspath{ {images/} }

\usepackage{caption}
\usepackage{subcaption}

\newcommand{\codebaselink}{\url{https://github.com/xuanrui-work/DeepLabelAlignment}}

\firstpageno{1}

\begin{document}

\title{
A Strategy for Label Alignment in Deep Neural Networks
}

\author{\name Xuanrui Zeng \email x64zeng@uwaterloo.ca \\
       \addr Eletrical and Computer Engineering\\
       University of Waterloo\\
       Waterloo, ON N2L 3G1, Canada}

\editor{Xuanrui Zeng}

\maketitle

\begin{abstract}
One recent research demonstrated successful application of the label alignment property for unsupervised domain adaptation in a linear regression settings. Instead of regularizing representation learning to be domain invariant, the research proposed to regularize the linear regression model to align with the top singular vectors of the data matrix from the target domain. In this work we expand upon this idea and generalize it to the case of deep learning, where we derive an alternative formulation of the original adaptation algorithm exploiting label alignment suitable for deep neural network. We also perform experiments to demonstrate that our approach achieves comparable performance to mainstream unsupervised domain adaptation methods while having stabler convergence. All experiments and implementations in our work can be found at the following codebase: \url{https://github.com/xuanrui-work/DeepLabelAlignment}.
\end{abstract} \label{abstract}

\begin{keywords}
Label Alignment, Neural Networks, Deep Learning
\end{keywords}

\section{Introduction}

Unsupervised domain adaptation is a subset of domain adaptation where the training data contains label for the source domain but not for the target domain. It is an inherently challenging problem in machine learning as ordinary models trained on the source domain aren't in anyway aware of the distribution difference between the source and target domain and don't have access to labeled target domain data for it to learn domain invariant representation.

As proposed by \cite{imani2022label}, a large proportion of binary classification and regression tasks exhibits the label alignment property, where the variation between the label and representation are mostly along the top principal components of the representation (\cite{imani2021understanding}). They further exploited this property to form a regularization objective on a linear regression setting and shown it to be feasible and effective for unsupervised domain adaptation.

In this work, we extend the work \cite{imani2022label} and intuitively deduce an alternative formulation to the label alignment objective proposed in their work tailored to deep neural networks (DNNs). We first build a proxy of the label alignment objective based on dimensionality reduction, we then exploit this proxy using an specially designed algorithm for DNNs, and lastly we empirically compare the performance of our method to 2 mainstream adversarial domain adaptation methods on the task of image classification to discuss its effectiveness and potential usage.

\section{Techniques}

\subsection{Previous Work from Imani et al. (2022)}
\cite{imani2022label} in their work deduced the following label alignment objective for linear regression settings in general:
    \begin{align}
    \min_{w}{\| \Phi w - y \|^2} &= \min_{w}{\| U \Sigma V^\top w - y \|^2} = \min_{w}{\| \Sigma V^\top w - U^\top y \|^2} \\
        &= \min_{w}{\sum_{i=1}^d (\sigma_i w^V_i - y^U_i)^2 + \sum_{i=d+1}^n (y^U_i)^2} = \min_{w}{\sum_{i=1}^d (\sigma_i w^V_i - y^U_i)^2} \label{eq:2} \\
        &= \min_{w}{\sum_{i=1}^k (\sigma_i w^V_i - y^U_i)^2 + \sum_{i=k+1}^d (\sigma_i w^V_i - y^U_i)^2} \label{eq:3} \\
        &= \min_{w}{\sum_{i=1}^k (\sigma_i w^V_i - y^U_i)^2 + \sum_{i=k+1}^d (\sigma_i w^V_i)^2} \label{eq:4}
    \end{align}
where $\Phi \in \mathbb{R}^{n \times d}$ is the representation matrix with each row being the features for the linear regression, $w \in \mathbb{R}^{d \times 1}$ is the weights of the linear regression, and $y \in \mathbb{R}^{n \times 1}$ is the label vector. And $\Phi = U \Sigma V^\top$ is the singular value decomposition (SVD) of $\Phi$: 
    \begin{align*}
    & U = [u_1, ..., u_n] \in  \mathbb{R}^{n \times n} &
    & V = [v_1, ..., v_d] \in  \mathbb{R}^{d \times d} &
    & \Sigma = diag([\sigma_1, ..., \sigma_d]) \in  \mathbb{R}^{n \times d} &
    \end{align*}
And the label alignment property: $y^U_i = 0, \forall i \in \{k+1, ..., d\}$, were used from \eqref{eq:2} to \eqref{eq:4}, assuming that the label alignment property holds for the first $k$ singular vectors.

In the original literature, the first term in \eqref{eq:4} was interpreted as linear regression on a smaller subspace of $\Phi$. While the second term in \eqref{eq:4} was called the \emph{label alignment regularization} and interpreted as minimizing $\sigma_i w^V_i = y^U_i, \forall i \in \{k+1, ..., d\}$, which has the effect of reducing the influence on the model's output from those singular vectors that are not the top principal components.

Based on the above interpretation, the following objective for unsupervised domain adaptation was further developed to adapt the linear regression model from a labeled source dataset $(\Phi, y)$ to an unlabeled target dataset $(\tilde{\Phi}, \tilde{y})$, with $\tilde{y}$ being unknown:
    \begin{align}
    \min_{w}{\| \Phi w - y \|^2 - \sum_{i=k+1}^d (\sigma_i w^V_i)^2 + \sum_{i=\tilde{k}+1}^d (\tilde{\sigma_i} w^{\tilde{V}}_i)^2} \label{eq:5}
    \end{align}
where the first term is the typical linear regression loss on the source domain, the second term removes the label alignment included in $\min_{w}{\| \Phi w - y \|^2}$ as shown in \eqref{eq:4}, and the third term enforces the label alignment on the target domain with rank $\tilde{k}$.

\subsection{Another Perspective}
Directly applying the same rigorous deduction above onto the case of deep neural networks (DNNs) is challenging due to both the diversity and non-linearity properties of DNNs. Instead, we start by reinterpreting the objective given by \eqref{eq:5} from a different perspective.

We start by combining \eqref{eq:5} with \eqref{eq:4} to form the following explicit objective equivalent to \eqref{eq:5}:
    \begin{align}
    & \min{\| \Phi w - y \|^2 - \sum_{i=k+1}^d (\sigma_i w^V_i)^2 + \sum_{i=\tilde{k}+1}^d (\tilde{\sigma_i} w^{\tilde{V}}_i)^2} \\
    &= \min{\sum_{i=1}^k (\sigma_i w^V_i - y^U_i)^2 + \sum_{i=k+1}^d (\sigma_i w^V_i)^2} - \sum_{i=k+1}^d (\sigma_i w^V_i)^2 + \sum_{i=\tilde{k}+1}^d (\tilde{\sigma_i} w^{\tilde{V}}_i)^2 \\
    &= \min{\sum_{i=1}^k (\sigma_i w^V_i - y^U_i)^2 + \sum_{i=\tilde{k}+1}^d (\tilde{\sigma_i} w^{\tilde{V}}_i)^2} \label{eq:8}
    \end{align}
We then make the assumption that the label alignment of the source and the target dataset have approximately the same rank, such that $\tilde{k} \approx k$. This assumption makes the two terms in objective \eqref{eq:8} independent, since then $\tilde{k} + 1 > k$ and all $w_i$ in the first term and all $w_i$ in the second term become mutually exclusive set. Under this assumption, \eqref{eq:8} can be decomposed into the following two respective objectives:
    \begin{numcases}{\eqref{eq:8} \equiv}
    & $\min_{w}{\sum_{i=1}^k (\sigma_i w^V_i - y^U_i)^2}$ \label{eq:9} \\ 
    & $\min_{w}{\sum_{i=k+1}^d (\tilde{\sigma_i} w^{\tilde{V}}_i)^2}$ \label{eq:10}
    \end{numcases}

We can rewrite objective \eqref{eq:9} and \eqref{eq:10} back into the following matrix forms respectively:
    \begin{align}
    \min_{w}{\sum_{i=1}^k (\sigma_i w^V_i - y^U_i)^2} &= \min_{w}{\| \Sigma^+ V^\top w - U^\top y \|^2} \\
    &= \min_{w}{\| U \Sigma^+ V^\top w - y \|^2} \\
    \min_{w}{\sum_{i=k+1}^d (\tilde{\sigma_i} w^{\tilde{V}}_i)^2} &= \min_{w}{\| \tilde{\Sigma}^- \tilde{V}^\top w \|^2} = \min_{w}{\| \tilde{U} \tilde{\Sigma}^- \tilde{V}^\top w \|^2} \\
    &= \min_{w}{\| \tilde{U} \tilde{\Sigma}^- \tilde{V}^\top w - y^o \|^2} \label{eq:14}
    \end{align}
where $U, V$ and $\tilde{U}, \tilde{V}$ follow from the SVD of $\Phi$ and $\tilde{\Phi}$ respectively, but $\Sigma^+$ is the reduced-\emph{upper} singular value matrix $\Sigma$ of $\Phi$ containing $\{\sigma_i | i \in \{1, ..., k\}\}$, and $\tilde{\Sigma}^-$ is the reduced-\emph{lower} singular value matrix $\tilde{\Sigma}$ of $\tilde{\Phi}$ containing $\{\tilde{\sigma_i} | i \in \{k+1, ..., d\}\}$. And a zero vector $y^o$ that doesn't affect the optimization is introduced at \eqref{eq:14}. More formally:
    \begin{align*}
    & \Sigma^+ = diag(\sigma_1, ..., \sigma_k, 0, ..., 0) \in \mathbb{R}^{n \times d} &
    & \tilde{\Sigma}^- = diag(0, ..., 0, \tilde{\sigma}_{k+1}, ..., \tilde{\sigma}_d) \in \mathbb{R}^{n \times d} &
    & y^o = \mathbf{0} &
    \end{align*}
Thus:
    \begin{numcases}{\eqref{eq:8} \equiv}
    & $\min_{w}{\| U \Sigma^+ V^\top w - y \|^2}$ \label{eq:15} \\ 
    & $\min_{w}{\| \tilde{U} \tilde{\Sigma}^- \tilde{V}^\top w - y^o \|^2}$ \label{eq:16}
    \end{numcases}

Objective \eqref{eq:15} can be interpreted as performing dimensionality reduction on $\Phi$ onto the top $k$ principal components, feeding the reduced $\Phi$ into the model, and minimzing the model's prediction loss on the reduced version of $\Phi$. Whereas objective \eqref{eq:16}, originally the label alignment regularization term, can be interpreted as performing dimensionality reduction on $\tilde{\Phi}$ onto the last $d - k$ principal components, feeding the reduced $\tilde{\Phi}$ into the model, and minimizing the model's output on the reduced version of $\tilde{\Phi}$.

\subsection{Onto Deep Neural Networks}
Following the intuition above, we can further deduce a general strategy for performing label alignment in DNNs. For demonstration, we start by discussing this part in the context of an example image classification task. Nevertheless, the same general strategy can be applied to other tasks as well.

Let's define $f: \hat{X} \rightarrow \hat{\Phi}$ to be a convolutional feature extractor (convolutional neural network), $g: \hat{\Phi} \rightarrow \hat{y}$ be a feedforward neural network, where $\hat{X} \in \mathbb{R}^{n \times c \times h \times w}$ is the input images in the form of a tensor, $\hat{\Phi} \in \mathbb{R}^{n \times d}$ is the flattened output feature map from the feature extractor, and $\hat{y} \in \mathbb{R}^{n \times m}$ is the output probability matrix with $m$ being the number of classes.

Let $(X, y)$ be the source dataset, ($\tilde{X}, \tilde{y}$) be the target dataset with $\tilde{y}$ unknown, $\Phi = f(X)$ be the feature map of $X$, $\tilde{\Phi} = f(\tilde{X})$ be that of $\tilde{X}$. Let $\Phi'(\Phi, k) = U \Sigma^+ V^\top$ be the reduced $\Phi$ and $\tilde{\Phi}'(\tilde{\Phi}, k) = \tilde{U} \tilde{\Sigma}^- \tilde{V}^\top$ be the reduced $\tilde{\Phi}$, using the same dimensionality reduction defined previously. To perform unsupervised domain adaptation using label alignment w.r.t. $\hat{\Phi}$, we transform and combine objective \eqref{eq:15} and \eqref{eq:16} to form the below objective function:
    \begin{align}
    & \min_{g}{\| g( U \Sigma^+ V^\top ) - y \|^2 + \lambda \| g( \tilde{U} \tilde{\Sigma}^- \tilde{V}^\top ) - y^o \|^2} \\
    &= \min_{g}{\| g(\Phi') - y \|^2 + \lambda \| g(\tilde{\Phi}') - y^o \|^2} \\
    &= \min_{f, g}{\| g[ \Phi'(f(X), k) ] - y \|^2 + \lambda \| g[ \tilde{\Phi}'(f(\tilde{X}), k) ] - y^o \|^2} \label{eq:19}
    \end{align}
where $\lambda$ is a hyperparameter controlling the strength of label alignment to the target domain, and we include $f$ in objective \eqref{eq:19} since we want to train the entire network $f(g(X))$ end-to-end.

Note that the first term in \eqref{eq:19} is simply the classification loss on the reduced $\Phi$ and is not limited to mean-squared-error loss. It can be replaced by other loss functions such as the cross-entropy loss if desired.

Also note that the dimentionality reduction on $\Phi$ and $\tilde{\Phi}$ depends on a suitable choice of $k$ for the construction of $\Sigma^+$ and $\tilde{\Sigma}^-$. In the original work of \cite{imani2022label}, $\Phi$ is a constant representation matrix irrespective of the optimization, and thus $k$ can be extracted by manually analyzing the principal components of $\Phi$. However in this case this is not feasible as $\Phi$ now varies according to $f$.

To address the above problem, we borrow some intuitions from \cite{imani2021understanding}. We make $k$ a variable and observe that the loss term in \eqref{eq:9} will be large if we choose $k >> k^*$ keeping all other terms constant, with $k^*$ being the theoretical optimal label alignment rank. Thus, following our previous derivations, minimizing the first term in \eqref{eq:19} w.r.t. $k$ only will have the effect of approximating $k \approx k^*$.

Expanding upon this idea, we make $k$ a learnable parameter for our optimization objective in \eqref{eq:19}. Furthermore, in practice in our experiment, we found insignificant performance difference when alternating the minimization of \eqref{eq:19} to be w.r.t. $f \& g$ and $k$ versus joint minimization of \eqref{eq:19} w.r.t. $f$, $g$, and $k$ all at once. Thus, we transform \eqref{eq:19} into the following final objective:
    \begin{align}
    \min_{f, g, k}{\| g[ \Phi'(f(X), k) ] - y \|^2 + \lambda \| g[ \tilde{\Phi}'(f(\tilde{X}), k) ] - y^o \|^2 + \gamma \| k \|^2} \label{eq:20}
    \end{align}
where the last term regularizes the learned $k$ to be small which is desired, and $\gamma$ is a hyperparameter controlling the weight of this regularization.

Additionally, to make \eqref{eq:20} differentiable w.r.t. $k$, we perform soft-gating on $\{\sigma_i | i \in \{1, ..., d\}\}$ and $\{\tilde{\sigma_i} | i \in \{1, ..., d\}\}$ using the sigmoid function to approximate selective indexing for the construction of $\Sigma^+$ and $\tilde{\Sigma}^-$:
    \begin{align*}
    & w_i = \frac{1}{1 + e^{\beta (i - k \cdot d)}} & \\
    & \Sigma^+ = diag(w_i \sigma_1, ..., w_d \sigma_d) \in \mathbb{R}^{n \times d} &
    & \tilde{\Sigma}^- = diag((1-w_i) \tilde{\sigma}_1, ..., (1-w_d) \tilde{\sigma}_d) \in \mathbb{R}^{n \times d} &
    \end{align*}
where $i \in [1, d]$ is the index of both $\sigma_i$ and $\tilde{\sigma_i}$, $k \in [0, 1]$ is our aforementioned $k$ but normalized, and $\beta > 0$ is a hyperparameter controlling the smoothness of the gating.

In practice, performing optimization of \eqref{eq:20} on large dataset is infeasible for DNNs, and batch optimization with a batch of data sampled from the dataset is used instead. To make our algorithm applicable to DNNs in general, we facilitate this pattern with objective \eqref{eq:20} with the assumption that the batch is large enough to be representative of our dataset.

Combining the aforementioned thoughts, Algorithm~\ref{alg:1} is the final resulted pseudocode encompassing our general strategy.
    \begin{algorithm}
    \caption{Unsupervised Domain Adaptation using Deep Label Alignment}\label{alg:1}
    \begin{algorithmic}
    \Require{\\
        hyperparameters $\lambda, \gamma, \beta$, learning rate $\alpha$, batch size $b$, iteration count $t$, \\
        source dataset $X$, target dataset $\tilde{X}$, source label $Y$ \\
        feature extractor network $f(\cdot)$, classification network $g(\cdot)$ \\
        classification loss function $cls\_loss(\cdot, \cdot)$ \\
    }
    \State{Initialize $f(\cdot), g(\cdot), \hat{k} \sim \mathcal{N}(0, 1)$}
    \For{$t$ iterations}
        \State{$(x, y), \tilde{x} \gets \text{sample batch with size $b$ from $(X, Y)$ and $\tilde{X}$}$}
        \State{$\Phi, \tilde{\Phi} \gets f(x), f(\tilde{x})$}
        \State{$(U, \Sigma, V), (\tilde{U}, \tilde{\Sigma}, \tilde{V}) \gets SVD(\Phi), SVD(\tilde{\Phi})$}
        \State{$k \gets sigmoid(\hat{k})$}
        \State{$\Sigma^+ \gets \text{construct $\Sigma^+$ using $\Sigma$ and $k$}$}
        \State{$\tilde{\Sigma}^- \gets \text{construct $\tilde{\Sigma}^-$ using $\tilde{\Sigma}$ and $k$}$}
        \State{$y^o \gets \mathbf{0}$}
        \State{Perform gradient step w.r.t. $cls\_loss[ g( U \Sigma^+ V^\top ), y ] + \lambda \| g( \tilde{U} \tilde{\Sigma}^- \tilde{V}^\top ) - y^o \|^2 + \gamma \| k \|^2$ with step-size $\alpha$, update $f(\cdot), g(\cdot), \hat{k}$}
    \EndFor
    \end{algorithmic}
    \end{algorithm}

\section{Evaluation} \label{evaluation}
In this section, we compare our approach to two mainstream approaches in unsupervised domain adaptation: Adversarial Discriminative Domain Adaptation (ADDA) by \cite{tzeng2017adversarial} and Domain-Adversarial Training of Neural Networks (DANN) by \cite{ganin2015unsupervised}. Both of which are domain adversarial based methods utilizing a domain classifier/discriminator with the goal of learning domain-invariant representations at the intermediate layers within a neural network.

\begin{figure}
    \centering
    \includegraphics[width=0.75\textwidth]{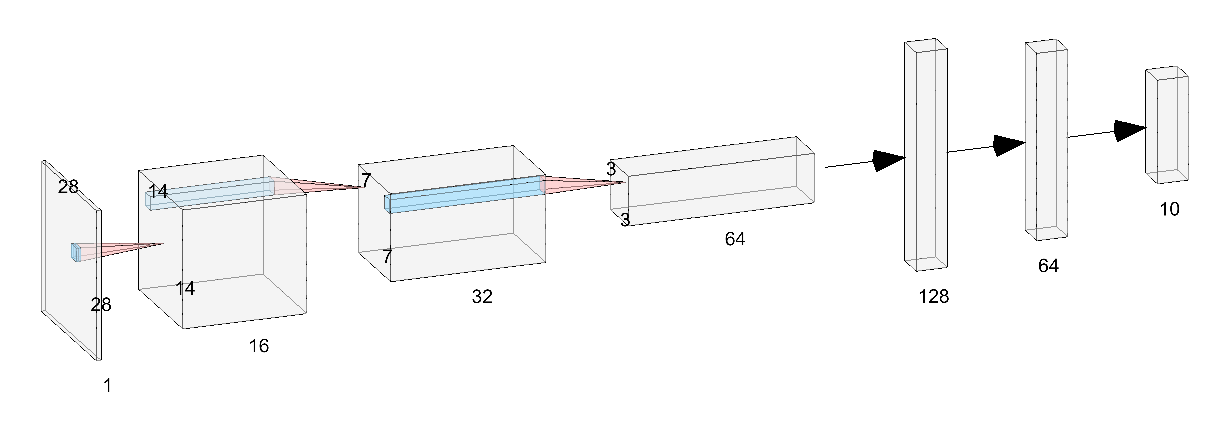}
    \caption{Architecture of the image classification network for our experiment.}
    \label{fig:1}
    \end{figure}

To carry out our comparison, we build a toy neural network with the architecture shown in Figure~\ref{fig:1} for image classification. We then perform unsupervised domain adaptation on the network using our method, ADDA, and DANN for MNIST $\rightarrow$ USPS, where MNIST is the labeled source dataset and USPS is the unlabeled target dataset, and we utilize the labels in the USPS for validation and testing only. We denote our method by DLA (Deep Label Alignment) for brevity.

\begin{figure}
    \centering
        \begin{subfigure}[b]{0.3\textwidth}
        \centering
        \includegraphics[width=\textwidth]{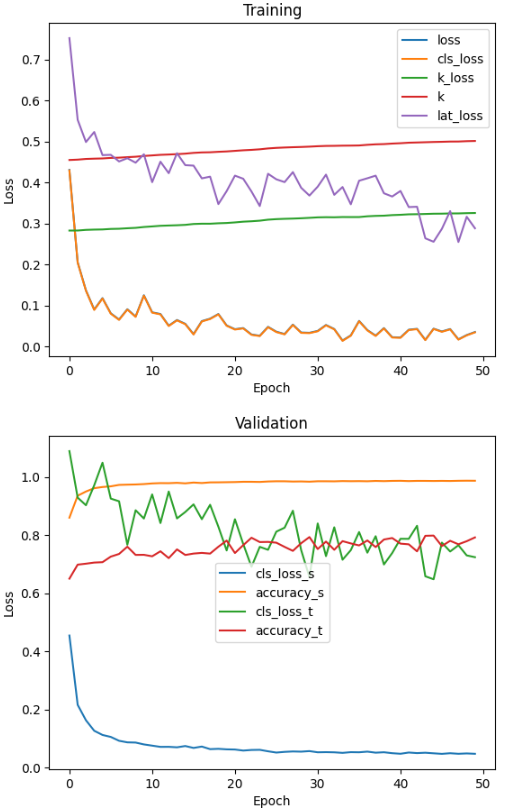}
        \caption{DLA}
        \label{fig:2:a}
        \end{subfigure}
        \hfill
        \begin{subfigure}[b]{0.3\textwidth}
        \centering
        \includegraphics[width=\textwidth]{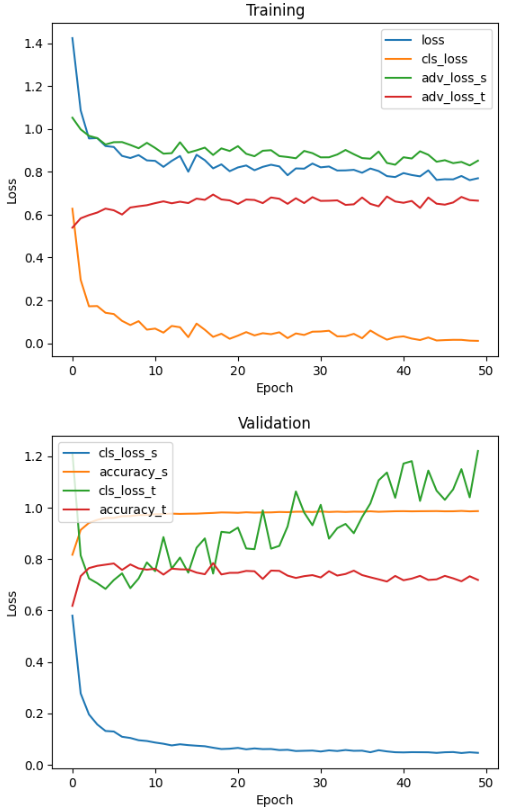}
        \caption{ADDA}
        \label{fig:2:b}
        \end{subfigure}
        \hfill
        \begin{subfigure}[b]{0.3\textwidth}
        \centering
        \includegraphics[width=\textwidth]{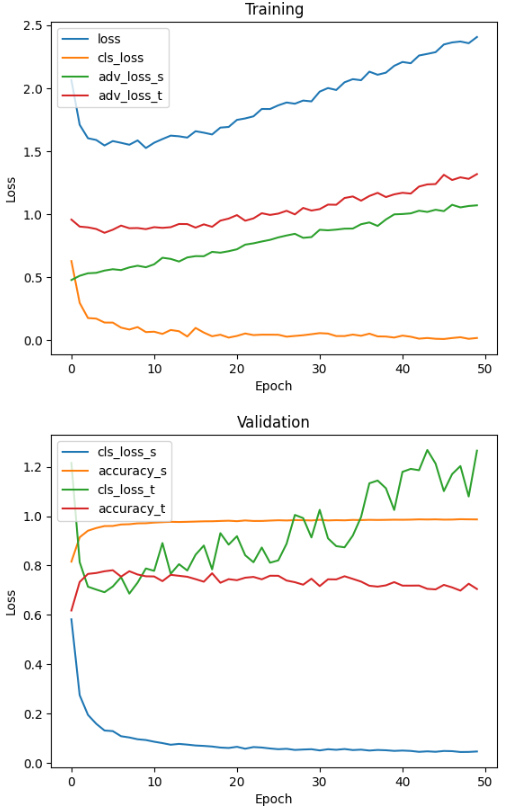}
        \caption{DANN}
        \label{fig:2:c}
        \end{subfigure}
    \caption{Training curves of: (a) our approach, (b) ADDA, and (c) DANN. Comparing to ADDA and DANN, our approach shows a stabler convergence for the classification loss in the target domain.}
    \label{fig:2}
    \end{figure}

Figure~\ref{fig:2} contains the training curves for the different methods and Table~\ref{tab:1} contains the final test accuracies, all averaged over 5 runs of each method. All methods are lightly tuned for good convergence over $\approx$2100 steps with a batch size of 128 and learning rate of $1e^{-3}$. Additionally, we use the following hyperparameters for our method: $\lambda=1e^{-3}, \gamma=1e^{-3}, \beta=5.0$. Hyperparameters for other methods can be found in our codebase at: \codebaselink.

Based on the results, we observe that our approach exhibits stabler training curves compared to ADDA and DANN while achieving a comparable accuracy. The instability in the training curves of ADDA and DANN is likely due to them utilizing adversarial training betweein the classifier network and the domain discriminator network, whereas in our approach the training curve is more stable as it utilizes the label alignment property instead.

    \begin{table}
    \centering
    \begin{tabular}{||c c c c||}
    \hline
    No Adaptation & DLA & ADDA & DANN \\
    \hline
    76.95 & \textbf{79.14} & 78.80 & 78.95 \\
    \hline
    \end{tabular}
    \caption{Test accuracies of our approach, ADDA, and DANN for MNIST $\rightarrow$ USPS tested on USPS test-set. Our approach achieved comparable accuracy on the target domain compared to both ADDA and DANN.}
    \label{tab:1}
    \end{table}

\section{Conclusion}
Based on our evaluation, we conclude that our extension to the work by \cite{imani2022label} onto deep neural networks is successful and that our approach is effective for unsupervised domain adaptation. In this work, we translated the core intuition behind label alignment and its objective into the language of deep learning and demonstrated its successful application in deep neural networks. For future research, we would recommend the following list of work given our current progress:
    \begin{enumerate}
        \item Our approach is based on intuitions and loose proofs. More rigorous proofs are needed to better understand the theories behind our approach and some of its theoretical properties.
        \item Our method relies on the assumption that the source and target dataset have approximately the same label alignment rank. This assumption needs further investigation and validation.
        \item We have only tested our proposed method on the adaptation of a single task, image classification, using only one dataset, the MNIST $\rightarrow$ USPS. Evaluating our method on the adaptation of different tasks with different datasets is desired to better compare our method with other mainstream methods.
        \item Interestingly, in our work we discovered that dropping the second term in Eq. \eqref{eq:20} to form the objective $$\min_{f, g, k}{\| g[ \Phi'(f(X), k) ] - y \|^2 + \gamma \| k \|^2}$$ has the effect of regularizing and preventing overfitting for supervised learning on the training dataset outside of the context of domain adaptation. We refer to this as the \emph{partial label alignment} objective and it can be further investigated to potentially identify another useful regularizer in addtion to the $l_1$ and $l_2$ regularizer.
    \end{enumerate}



\newpage

\appendix

\vskip 0.2in
\bibliography{references}

\end{document}